# A Deep Learning Approach to Predict Hamburg Rutting Curve

**Hamed Majidifard**
Ph.D. Candidate
Civil and Environmental Engineering Department
University of Missouri-Columbia
Email: hmhtb@mail.missouri.edu
*Corresponding author

**Behnam Jahangiri**
Ph.D. Candidate
Civil and Environmental Engineering Department
University of Missouri-Columbia
Email: bjctn@mail.missouri.edu

**Punyaslok Rath**
Ph.D. Candidate
Civil and Environmental Engineering Department
University of Missouri-Columbia
Email: prath@mail.missouri.edu

**Amir H. Alavi**
Assistant Professor
Department of Civil and Environmental Engineering
University of Pittsburgh
Email: alavi@pitt.edu

**William G. Buttlar**
Professor and Glen Barton Chair in Flexible Pavements
Civil and Environmental Engineering Department
University of Missouri-Columbia
Email: buttlarw@missouri.edu

**Abstract**
Rutting continues to be one of the principal distresses in asphalt pavements worldwide. This type of distress is caused by permanent deformation and shear failure of the asphalt mix under the repetition of heavy loads. The Hamburg wheel tracking test (HWTT) is a widely used testing procedure designed to accelerate, and to simulate the rutting phenomena in the laboratory. Rut depth, as one of the outputs of the HWTT, is dependent on a number of parameters related to mix design and testing conditions. This study introduces a new model for predicting the rutting depth of asphalt mixtures using a deep learning technique - the convolution neural network (CNN). A database containing a comprehensive collection of HWTT results was used to develop a CNN-based machine learning prediction model. The database includes 10,000 rutting depth data points measured across a large variety of asphalt mixtures. The model has been formulated in terms of known influencing mixture variables such as asphalt binder high temperature performance grade, mixture type, aggregate size, aggregate gradation, asphalt content, total asphalt binder recycling content, and testing parameters, including testing temperature and number of wheel passes. A rigorous validation process was used to assess the accuracy of the model to predict total rut depth and the HWTT rutting curve. A sensitivity analysis is presented, which evaluates the effect of the investigated variables on rutting depth predictions by the CNN model. The model can be used as a tool to estimate the rut depth in asphalt mixtures when laboratory testing is not feasible, or for cost saving, pre-design trials.

# 1. INTRODUCTION

Rutting is one of the major distresses in the asphalt pavements compromising the ride quality and safety of roads. This type of distress is caused by permanent deformation and shear failure of the asphalt mix under the repetition of heavy loads in during warm and/or wet seasons. This phenomenon is especially prevalent in hot climates and on heavily trafficked routes. In order to characterize the rutting process, numerous experimental studies have been conducted [1]–[3]. Among the implemented tests, the Hamburg wheel tracking test (HWTT) has gained considerable attention from the researchers and agencies across the United States [4]. This test applies thousands of wheel load passes to simulate the traffic loads on hot mix asphalt (HMA) specimens in an accelerated fashion at a selected temperature, such as 50 °C, and typically using a temperature controlled water bath as the heating medium. The HWTT is performed in accordance with the AASHTO T324 standard. Rut depth along the sample as a function of wheel pass represents the main output of the HWTT. The averaged rut depth versus wheel pass is dependent on a number of factors related to the mix design and test conditions. Recently, Lv et al. [5] used the HWTT to evaluate the high temperature performance of asphalt mixes modified by styrene-butadiene-styrene (SBS), polyphosphoric acid, and rubber. To this end, different dosages of additives and a base binder of PG 64-22 were used to make the binder system. The results showed that the rut depth decreased as a result of SBS-polymer modification. However, increasing the dosage did not necessarily result in a reduction of permanent deformation. Therefore, an optimum additive content was identified for each additive. In another study, Ziari et al. [6] added 5, 10, and 15 % of minus-40 mesh rubber to a dense-graded control mix. The rut depth under the Hamburg test was reduced by up to 40% at high rubber dosages. Vahidi et al. [7] tested a mixture that had 40% Reclaimed Asphalt Pavement (RAP) content as the control mix. It was reported that adding 10% of ground tire rubber (GTR) by weight of binder to this mix significantly lowered the Hamburg rut depth [7]. This observation was attributed to the stiffened binder resulting from the absorption of volatiles in the binder by the rubber particles, and due to the stiffness of the swelled rubber [7].

Although most of agencies such as state departments of transportation (DOTs) across the United States conduct the Hamburg test at 50 °C, some researchers have linked the recommended testing temperature to the plan binder grade required for a project in a given climate under a given traffic level. Kim et al. [8] have studied the effect of temperature on rut depth at 50, 64, and 70 °C in the state of Georgia. Walubita *et al*. [9]attributed the rutting failures observed in Texas to the low Hamburg temperature compared to the actual temperature that the pavement experiences during its service life in the southern part of the United States. Conducting the Hamburg test at higher temperatures such as 60 and 70 °C could more accurately screen rutting prone mixtures [9]. In colder climates such as Wisconsin, lower temperatures have been used to evaluate the rutting resistance of the mixtures. Swiertz et al. [10] performed the Hamburg test at 45 and 50 °C and observed that the number of passes to reach 12.5 mm of rut depth at 50 °C could be as high as three times higher than that at 45 °C. Considering four nominal maximum aggregate sizes (NMAS) including 25, 19, 12.5, 9.5, and 4.75 mm of Superpave mixtures, Swiertz et al. [10] investigated the effect of aggregate gradation on the rut depth. The Hamburg testing results at 50 °C under 20,000 wheel passes showed that the rut depth slightly increased as the NMAS increased from 4.75 to 25 mm. Accordingly, better rutting performance for coarser gradations was reported by Larrian and Tarefder [11]. Batioja et al. [12] studied the potential of Hamburg as a quality assurance test to be implemented in the state of Indiana. To this end, the effects of binder grade, NMAS, and testing temperature on both field cores and plant produced lab compacted samples were investigated through the rutting resistance index (RRI). The results indicated a clear distinction of rut depth between the mixtures with different binder grades. The testing results showed that Hamburg testing could properly capture the effect of aggregate type, with softer aggregates leading to higher rut depths in the HWTT [13].

In an attempt to introduce more parameters related to moisture and rutting susceptibility of the mixtures, Yin et al. [14] showed that incorporating more recycled materials results in a better rutting resistance, yet increases stripping potential. Accordingly, experimental results in [15] showed that adding 50% of RAP could reduce the rut depth of a virgin mix from 2.83 to 1.19 mm at 10,000 passes. Ozer et al., [16] added 2.5, 5, and 7.5 % recycled asphalt shingles (RAS) by weight of the mix and observed a significant drop in rut depth in the Hamburg test as the recycled material content increased. The beneficial effect of recycled materials on high temperature performance has also been reported by other researchers [17], [18][19][20]–[22]. As an example, Jahangiri et al. studied eighteen mixtures used to pave the roads across Missouri. The measured rut depth under Hamburg results revealed the positive effect of stiffer binder system as a result of higher asphalt binder replacement by recycled materials, including RAP and RAS [23].

Buss et al. [24] compared laboratory compacted samples to the field cores in terms of Hamburg rut depth. To this end, the asphalt materials used to pave four sections were sampled, and laboratory specimens were fabricated. In addition, the field cores were taken from the sections and tested in the Hamburg device at 50 °C. It was observed that in all four mixtures, the field cores accumulated lower rut depth as compared to the gyratory compacted samples in the lab. However, results reported in [12] showed better rutting resistance in plant-produced, laboratory-compacted (PPLC) specimens. The significant effect of sample type such as field core, PPLC and lab-produced, lab-compacted (LPLC) has been reported in other studies [25].

Recently, there has been a growing trend withing the pavement community to deploy advanced machine learning and deep learning methods for the characterization of asphalt pavements. In this arena, Majidifard et al. [26], [27] used genetic expression programming (GEP) to develop prediction models for both fracture energy of asphalt at low temperature and rutting performance at high temperature. This study proposes a new deep learning approach to predict the rutting depth versus wheel pass curve obtained from the HWTT as a function of asphalt mixture properties. Considering the increasing use of the HWTT in the balanced mix design movement across the United States [11], [14], [24], [28], such models can provide both time and cost savings for the purposes of initial design and material screening. A subset of deep learning entitled the convolution neural network (CNN) method is used herein to create an analytical model that can be easily programmed into a spreadsheet or into cloud-based services. Sensitivity and parametric analyses are then performed to validate the reliability of the developed machine learning model.

## 2. CONVOLUTION NEURAL NETWORK

CNNs is a branch of deep learning motivated by physiological processes between neurons that mimic visual cortex systems [29]. A CNN structure is created by an input, multiple hidden layers, and an output layer. The hidden layers typically include a series of convolutional layers that convolve with multiplication or other dot products. Each hidden layer is followed by an activation function, which is commonly taken as Rectified linear units (RELU) layer. Hidden layers are consisting of neurons with learnable weights and biases. Each neuron takes various inputs, exerts a weighted sum across them, transfers it within an activation function, and returns an output. A loss function at the end of the structure can minimize the error and re-assign new weights to the neurons. Multilayer perceptrons are regulated to form a CNN. In a multilayer perceptrons structure, the layers are fully connected. Thus, each neuron in one layer is connected to all neurons in the next layer [30]–[33].

The application of CNN in the pavement arena generally involves either classification or regression approaches. Most classification applications using the CNN model involve detecting and classifying pavement distresses [34]–[36]. For instance, Zhang et al. [37] established CrackNet software using CNN to detect patches [37]. Also, Majidifard et al. [35] developed YOLO v2 and Faster R-CNN frameworks to classify pavement distress using a comprehensive database. Also, Majidifard et al. [34] developed an automatic pavement distress detection tool to classify and quantify the severity of the distresses via hybrid YOLO and U-Net models [34]. Jiang et al. [38] used CNN to classify asphalt mixtures components (aggregate, mastic, and air void) from X-ray images [38]. There are limited studies that focus on using CNNs for regression problems. Gong et al. [39] have recently used a deep neural network to predict pavement rutting for the mechanistic-empirical pavement design guide [39]. The current study aims at harnessing the power of CNNs for developing HWTT rutting prediction models.

## 3. EXPERIMENTAL DATABASE

A variety of testing specimens fabricated from different asphalt mixtures, including plant-produced/laboratory-compacted ('Plant') mixtures, and laboratory-produced laboratory-compacted ('Lab') mixtures were tested at different temperatures. The aggregate and binder were mixed together in a bucket at the appropriate, mix-specific mixing temperature and kept in the oven for 2 hours at the corresponding, appropriate compaction temperature according to AASHTO R35. Mixing and compaction temperatures varied based on the Superpave performance grade of the binder and the resulting viscosity across the production temperature range. For plant mixtures, loose mixture was heated in an oven until the compaction temperature was reached. Afterward, all samples were compacted to 7±0.5% air void. Details about the mixtures investigated are provided in Table 1, with additional details provided elsewhere [23], [40]–[42][43].

**Table 1.** Range of materials [a] and other mixture characteristics represented in the 29 asphalt sections evaluated in this study

| Mixture Type | Virgin Binder Grades in Study Mixes | Asphalt Content Range (%) | Gradation Types | NMAS (mm) Range | ABR (%) Range | Additives Used |
|---|---|---|---|---|---|---|
| Plant (17 Mix), Lab (12 Mix) | PG 76-22 PG 70-22 PG 70-28 PG 64-22 V PG 64-22 H PG 64-22 PG 64-34 PG 58-28 PG 46-34 | 5.1 to 7.9 | Dense (23 Mix) SMA (6 Mix) | 19 (2 Mix) 12.5 (20 Mix), 9.5 (4 Mix), 4.75 (3 Mix) | 0 to 48.4 | 1. Anti-stripping agent ('Morelife T280') 2. Warm-mix additive ('Evotherm') 3. Rejuvenator additive ('EvoFlex CA') 4. Anti-stripping agent ('IPC-70') 5. Warm-mix additive ('PC 2106') 6. Crumb rubber (dry process and terminal blend) 7. Steel Fiber |

[a] For convenience, property ranges are given. Details are provided in [23], [40]–[42][43].

On the basis of the literature reviewed, this study presents a new CNN-based prediction model for the HWTT rut depth vs. wheel pass as follows:

$$Rut\ depth\ (mm) = f(Mix\ type, UTI, HTPG, AC, NMAS, ABR, RAP, RAS, G, AT, CRC, T, Pass) \quad (1)$$

where,

*Mix type:* Plant or lab-compacted mixture (Plant (loos asphalt mixture collected from plant and compacted in the lab) = 1, Lab (produced from virgin materials in lab) = 2)

*UTI* (°C): Useful temperature interval (i.e., for PG 64-22 binder, UTI = 64-(-22) = 86°C)

*HTPG* (°C): High temperature performance grade (for PG 64-22 binder, this is 64°C)

*AC* (%): Asphalt content

*NMAS* (mm): Nominal maximum aggregate size. In this study, NMAS can be either 4.75mm, 9.5mm, 12.5 mm or 19 mm

*ABR (%):* Asphalt Binder Replacement (or ABR, ratio of recycled binder to total binder)

*RAP* (%): Reclaimed asphalt pavement, % by ABR

*RAS* (%): Reclaimed asphalt shingles, % by ABR

*G*: Gradation type (Dense: 1; SMA: 2)

*AT*: Aggregate Type (Limestone: 1; Granite: 2)

*CRC* (%): Crumb rubber content (rubber percentage in virgin binder)

*T* (°C): Testing temperature

*Pass*: Number of wheel passes

The models were developed using 10,000 data points from HWTT results from various tests conducted on 29 materials from pavements in mid-continental United States. Fifty tests were performed according to AASHTO-T324 [44]. For each test, three samples were tested, and averaged results were used. Therefore, 150 samples were tested to collect 10,000 test results. As shown in Figure 1, the 71.7 kg loaded steel wheel runs over the samples placed in a water bath at 50 °C (Figure 1). The vertical deformation of the specimen is recorded along with the number of wheel passes. The test ends when either the specimen deforms more than 20 mm or the number of passes exceeds 20,000 (Figure 2). Each curve contained 200 data points (Figure 2). Moreover, this study evaluated different test temperatures for the Hamburg wheel test. Various DOTs and agencies are adopting Hamburg test temperatures according to the prevailing climatic conditions instead of the standardized recommended value of 50 °C. Therefore, the proposed model can be considered a practical tool for design purposes by DOTs as it incorporates the temperature as a variable in its predictions.

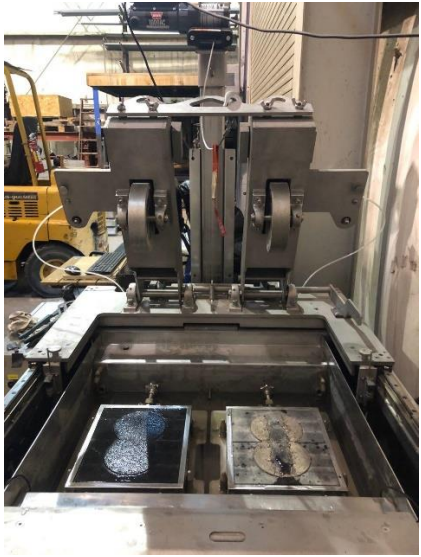
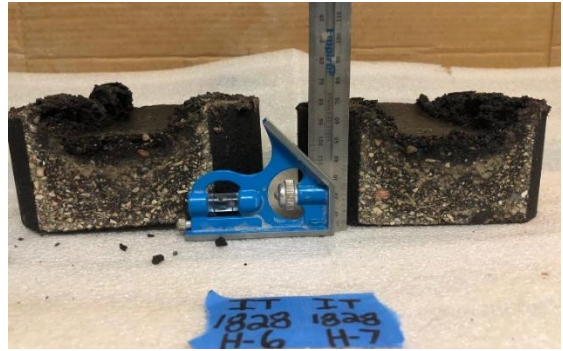

(a)                          (b)

**Figure 1.** (a) A Hamburg wheel tracking device, (b) Asphalt sample after 20,000 passes in the Hamburg device.

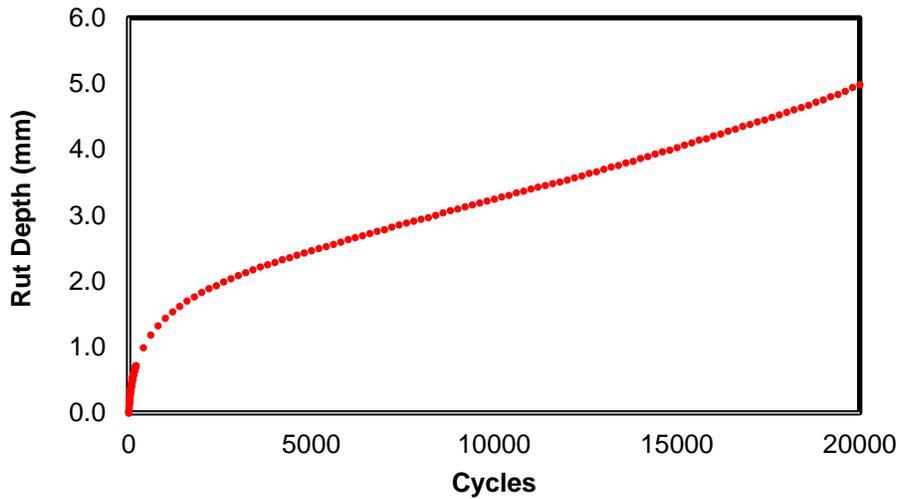

**Figure 2.** Typical rut depth curve versus number of passes in Hamburg wheel tracking test (200 data points).

Ranges of the parameters used to develop the prediction model are summarized in Table 2.

Table 2. Descriptive statistics of the variables

|  | Mean | Median | Mode | Max | Min | Range |
|---|---|---|---|---|---|---|
| **Mix type (Plant=1, Lab=2)** | 1.4 | 1 | 1 | 2 | 1 | 1 |
| UTI | 86.4 | 86 | 86 | 98 | 80 | 18 |
| High PG | 56.8 | 58 | 58 | 70 | 46 | 24 |
| Asphalt Content (AC) (%) | 5.7 | 5.4 | 5.1 | 7.9 | 5.1 | 2.8 |
| ABR (%) | 31.2 | 32.5 | 35.2 | 48.4 | 17 | 31.4 |
| NMAS (mm) | 11.8 | 12.5 | 12.5 | 19 | 4.8 | 14.3 |
| RAP (%) | 24.6 | 24 | 35.2 | 35.3 | 0 | 35.3 |
| RAS (%) | 6.6 | 0 | 0 | 33 | 0 | 33 |
| Gradation (Dense=1, SMA=2) | 1.2 | 1 | 1 | 2 | 1 | 1 |
| Agg Type (limestone=1, Granite=2) | 1.2 | 1 | 1 | 2 | 1 | 1 |
| Crumb Rubber Content (CRC) | 3.7 | 0 | 0 | 20 | 0 | 20 |
| Temp (°C) | 51.6 | 50 | 50 | 64 | 40 | 24 |
| Number of Passes | 5046.3 | 198 | 0 | 20000 | 0 | 20000 |
| Rut Depth (mm) | 1.6 | 0.8 | 0 | 19.8 | 0 | 19.8 |

Frequency histograms were formed to present the variability of the parameters (see Figure 3). As seen in the figure, the distribution of the variables is not uniform.

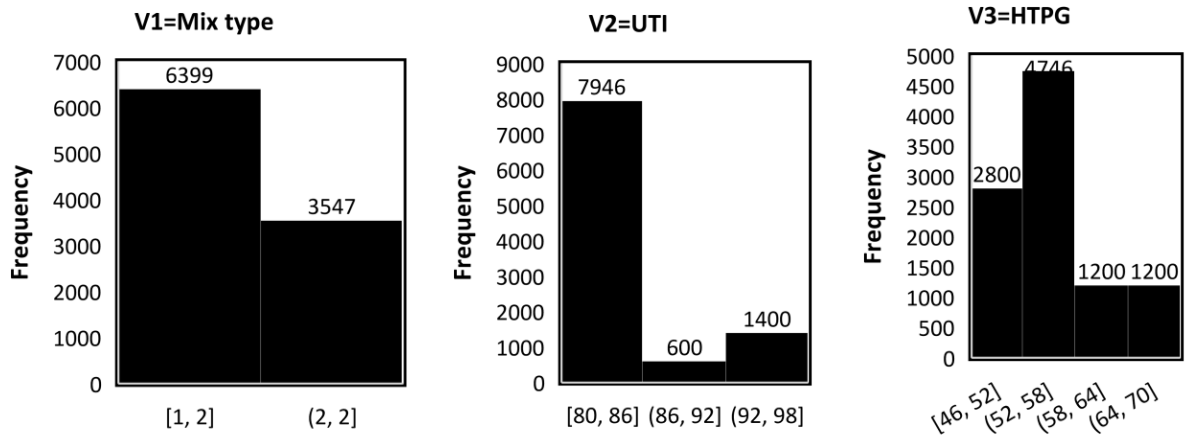

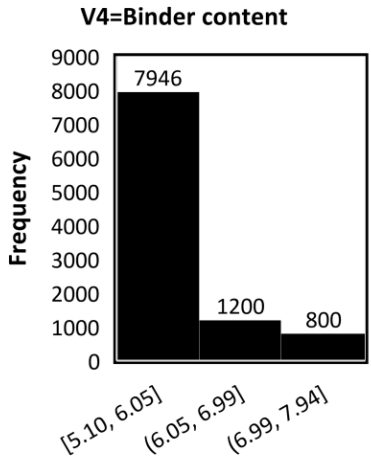
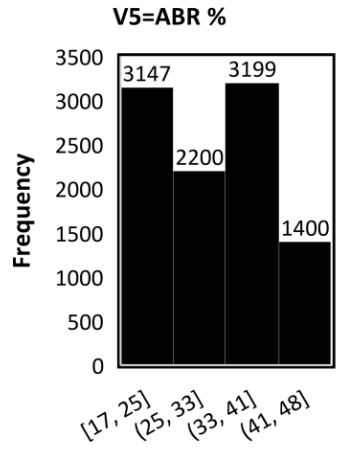
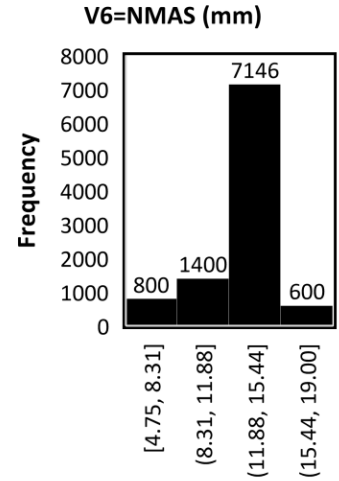
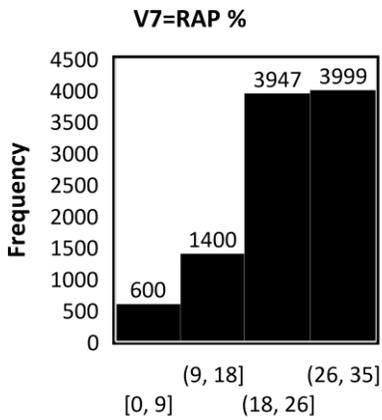
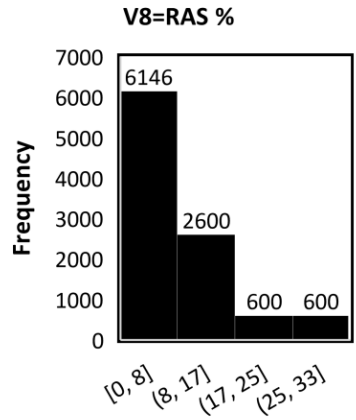
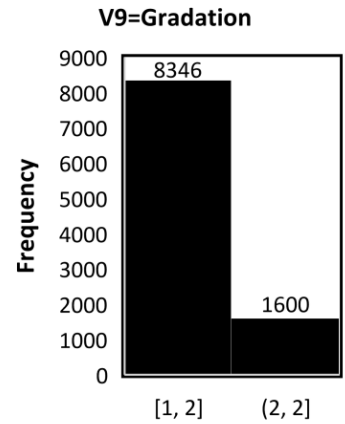
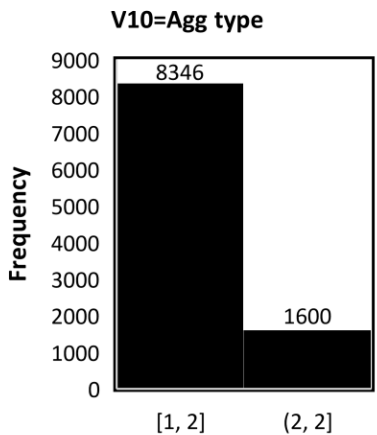
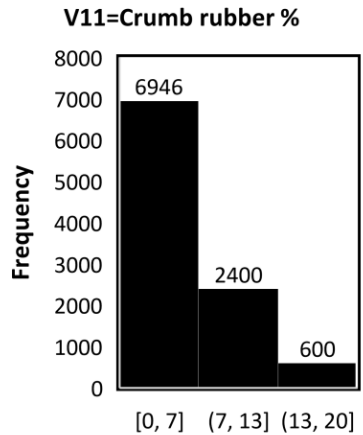
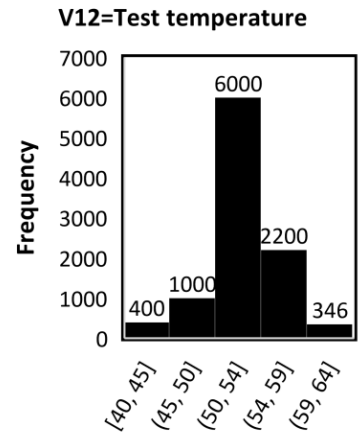

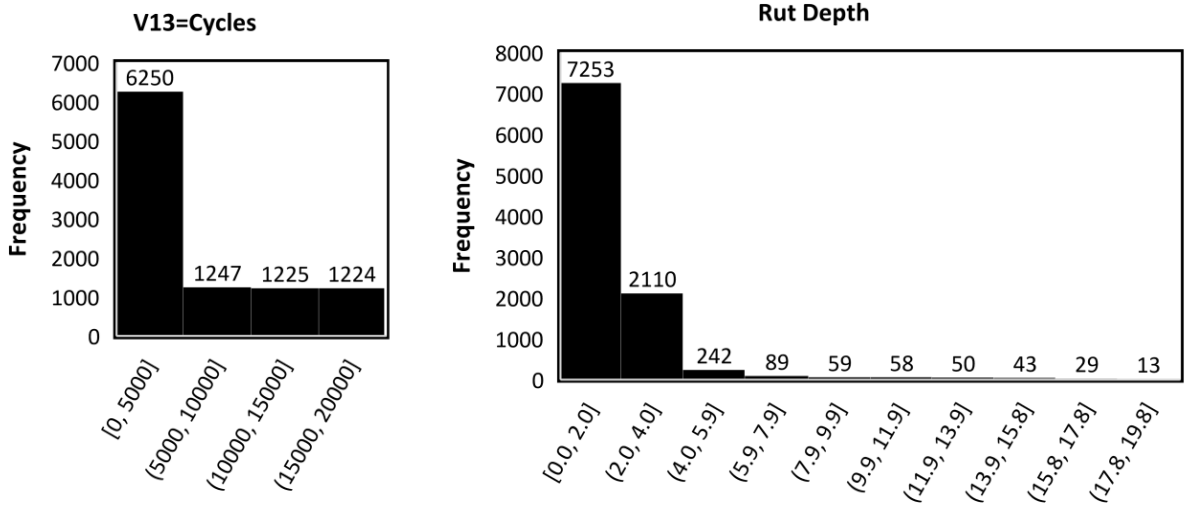

**Figure 3.** Distribution histograms of the variables used in the study

To avoid overfitting, the database was divided into three subsets:

- Learning: Samples of data used to fit the model (~ 70%).
- Testing: Samples of data used to provide an unbiased evaluation of a model fit on the training dataset while tuning model parameters (~ 15%).
- Validation: Sample of data used to provide an unbiased evaluation of the final model on unseen dataset (~ 15%).

Model calibration and preliminary evaluation were done using the learning and testing sets, respectively. The accuracy of the models was assessed using the following indexes:

- $R^2$: Correlation coefficient
- RMSE Root mean squared error
- MAE: Mean absolute error

$$R^2 = \frac{(\sum_{i=1}^{n}(O_i-\overline{O_i})\,(t_i-\overline{t_i}))^2}{\sum_{i=1}^{n}(O_i-\overline{O_i})^2 \sum_{i=1}^{n}(t_i-\overline{t_i})^2} \tag{2}$$

$$RMSE = \sqrt{\frac{\sum_{i=1}^{n}(O_i-t_i)^2}{n}} \tag{3}$$

$$MAE = \frac{\sum_{i=1}^{n}|O_i-t_i|}{n} \tag{4}$$

where,

$O_i$: Measured output

$t_i$: Predicted output

$\overline{O_i}$: Average of measured outputs

$\overline{t_i}$: Average of predicted outputs

$n$: Number of samples

## 4. DEVELOPMENT OF HWTT PREDICTION MODEL

### 4.1. Data Normalization

Feature normalization is the first step in training the dataset, which makes the training process faster and more accurate. Therefore, both the training and testing dataset were normalized, as normalization should be applied to any data used to train or test the model. Data normalization eliminates the measurement units for data, enabling the comparison of data from different sources. Normalization transforms data to have a mean of zero and a standard deviation of 1. This standardization is called a z-score, and data points can be standardized with the following formula:

$$Z_i = \frac{x_i - \bar{x}}{S} \tag{5}$$

Where:

$Z_i$ = Z-score-normalized data

$x_i$ is a data point $(x_1, x_2 \ldots x_n)$.

$\bar{x}$ is the sample mean.

S is the sample standard deviation.

### 4.2. Optimal CNN Structure

#### 4.2.1. Convolutional layer

The principal benefit of CNN compared to neural networks is the use of parameter distribution in the convolutional layer. The convolutional layer utilizes multiple filters to produce outputs instead of directly connecting each neuron set to the following layer [45].

*4.2.2. Activation units*

An activation function calculates the weighted sum and adds in bias to determine if a neuron should be activated or not [31]. A neural network without an activation function is basically just a linear regression model. The activation function applies the non-linear conversion to the input, initiating the capability to learn and produce more complex tasks. In neural networks, the weights and biases of the neurons are updated according to the output error. Activation functions make the back-propagation feasible since the gradients are supplied along with the error to renew the weights and biases. Rectifier Linear Unit (ReLU) is the most popular and efficient activation function recently [46][31] (Figure 4).

$$f(x) = \max(0, x) \tag{6}$$

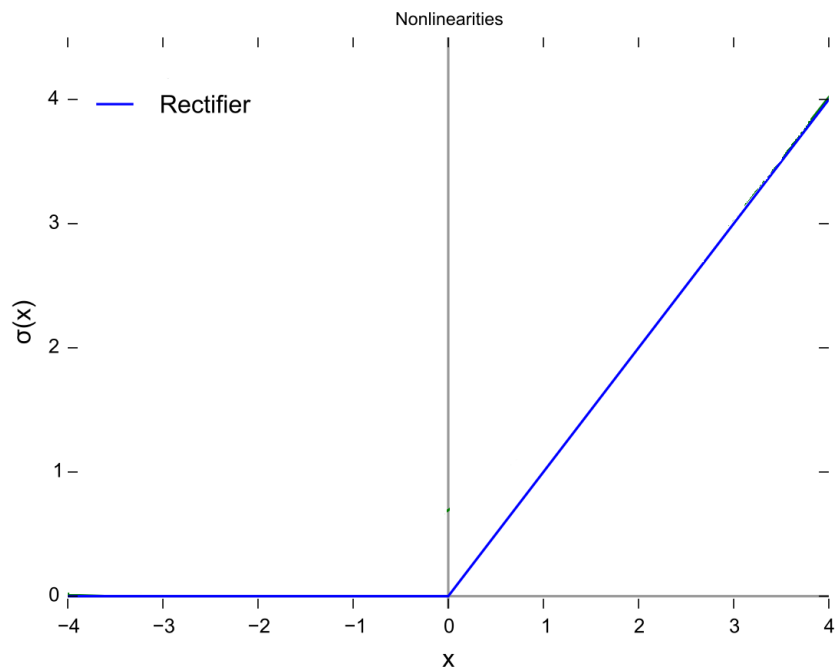

**Figure 4.** ReLU function [46]

ReLU is employed broadly in most deep learning models and greatly enhances model performance in most instances. Since the function derivation is "1" for positive values, it flags the errors to be back-propagated efficiently to avoid losing gradient. Also, the cost of computation reduces, and the model converges more accurately due to the simplicity of the function.

*4.2.3. Optimizing function*

The optimization technique is generally used in deep neural network training to minimize the loss function. The optimization function herein applied is the Root Mean Square Propagation (RMSProp) [47]. The default learning rate (0.001) suggested by the RMSprop was employed as an adaptive learning rate algorithm.

*4.2.4. Regularization of overfitting*

Regularization procedures are explicitly designed to prevent overfitting and consequently enhance generalization. Some regularization technique examples include: early stopping, weight decay through L2 or L1 parameter regularization, dropout, and dataset increment [30], [48], [49]. In this study, we used an early stopping callback that examines training condition for each epoch. If a considered number of epochs lapses without showing improvement, then the algorithm stops the training.

In this study, the proposed model consisted of a sequential model with two densely connected hidden layers and an output layer that delivered unique output values (Figure 5).

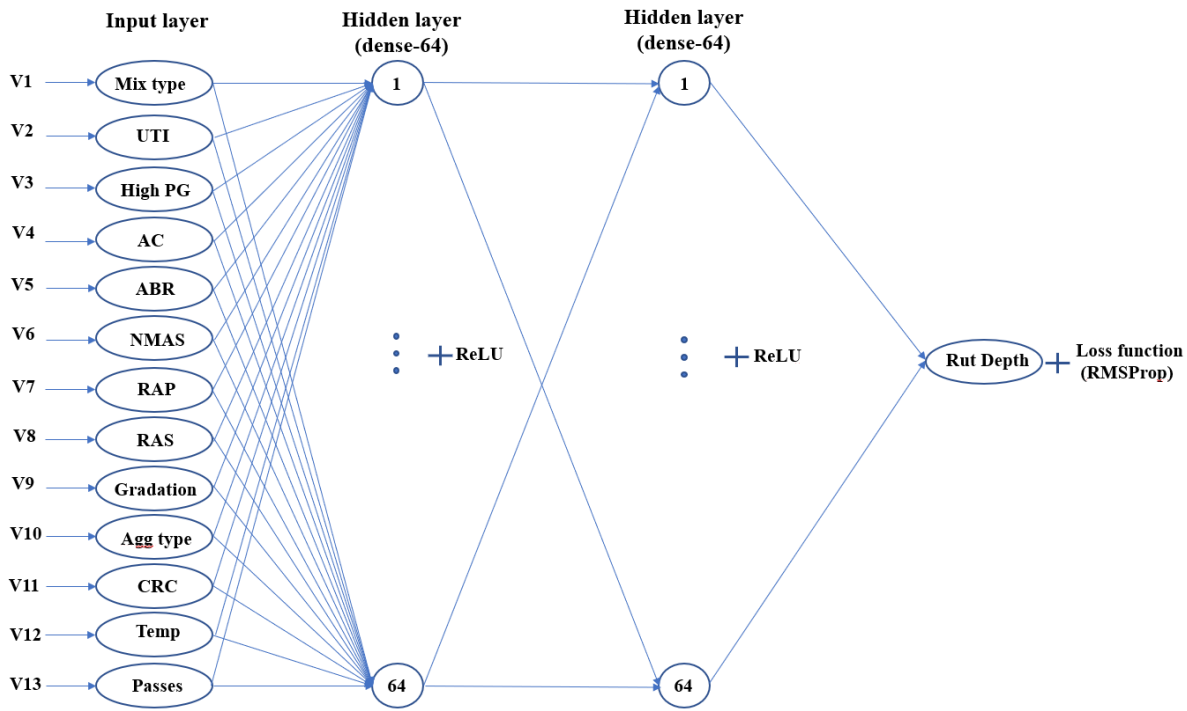

**Figure 5.** The optimal CNN structure

## 4. RESULTS

The framework was developed via Tensorflow and Keras libraries in python. A batch size of 10 was selected. The model was trained for 1000 epochs. The results for training and testing datasets were monitored (Figure 6). $R^2$ calculated for both the training and testing dataset are shown in Figure 8.

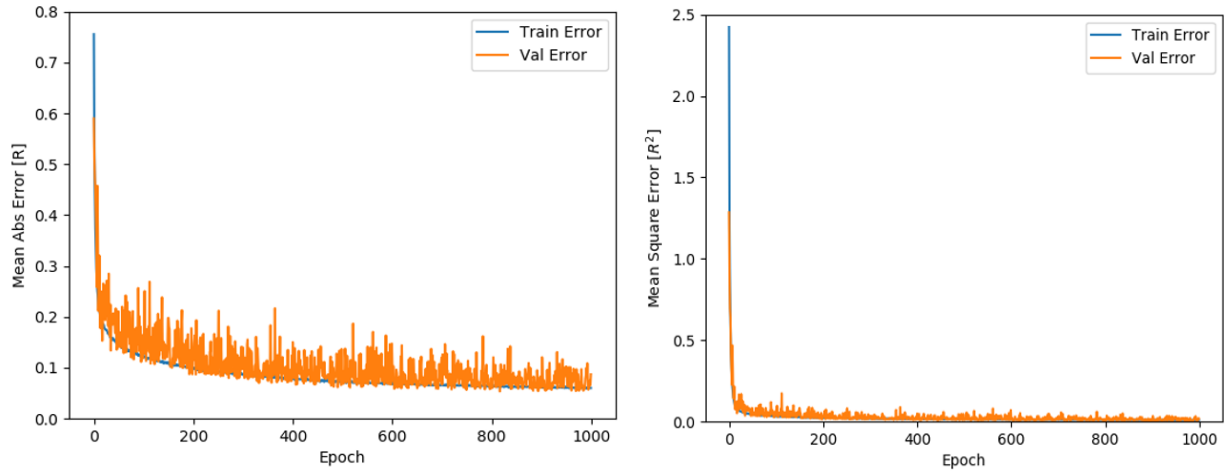

**Figure 6.** Performance of the calibrated model on the training and validation datasets for 1000 epochs.

As mentioned before, early stopping is a beneficial technique to avoid overfitting. Figure 7 shows insignificant improvement in the error after about 40 epochs.

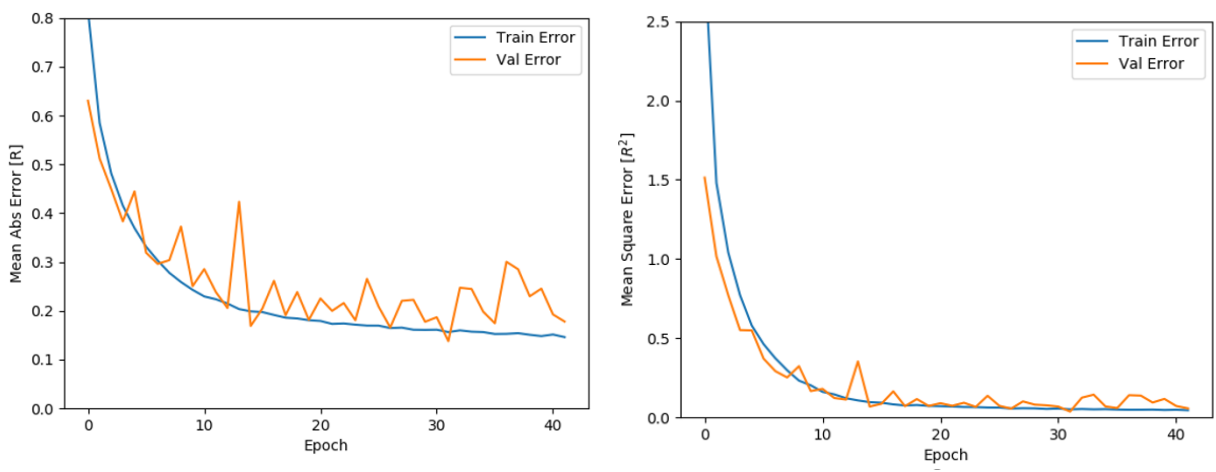

**Figure 7.** Performance of the calibrated model on the training and validation datasets for 40 epochs.

Figure 8 shows the performance indices of the optimal CNN model on the training, validation, and testing dataset, respectively. As seen, the proposed CNN model can predict the rut depth accurately.

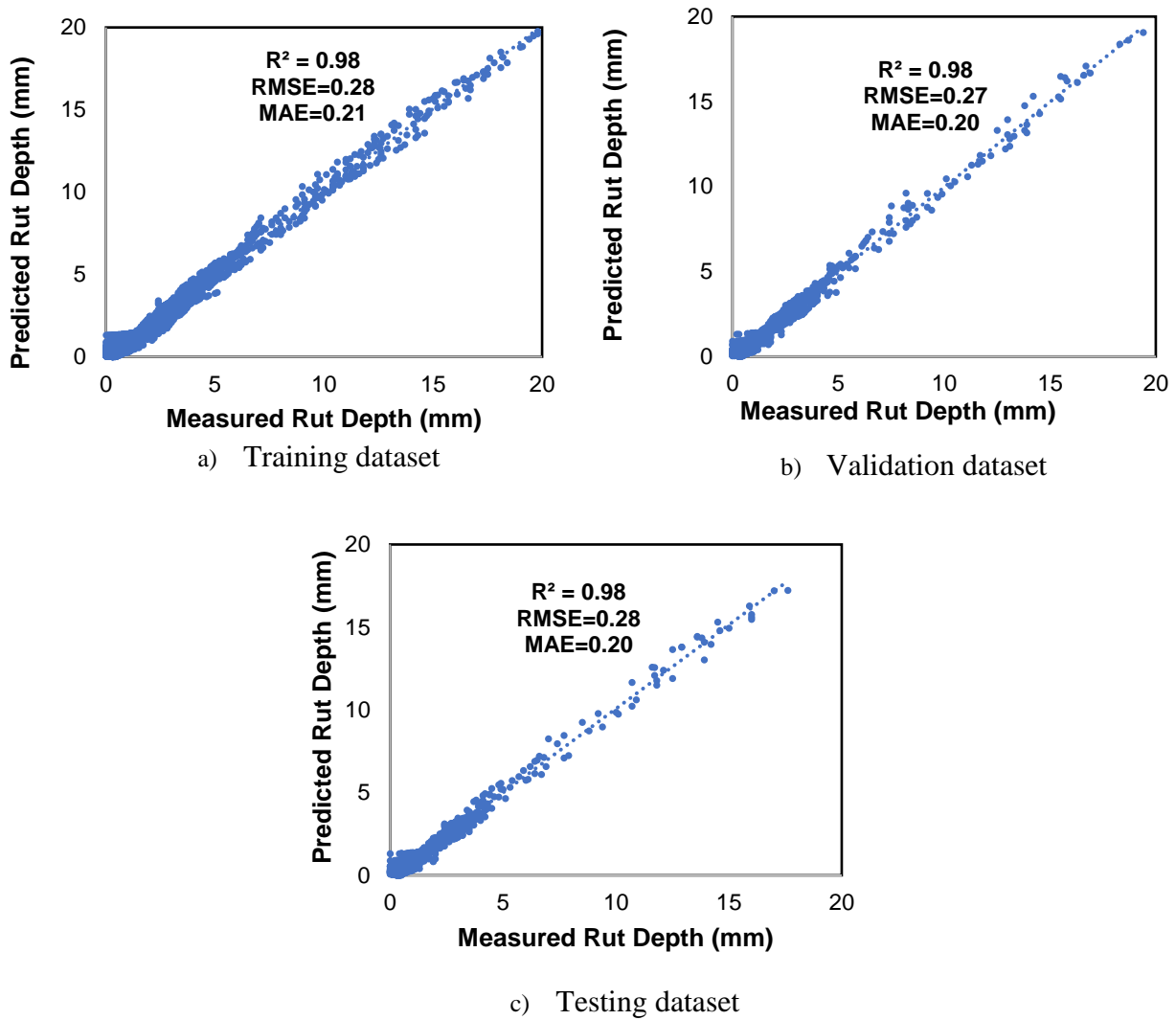

**Figure 8.** Measured versus predicted rut depth using the optimal CNN model: (a) training dataset, (b) validation dataset, and c) testing dataset.

## 5. SENSITIVITY ANALYSIS

The effectiveness of the predictive models can be evaluated by inspecting model sensitivity to known influencing factors. The CNN model presented in this study has been used to predict HWTT results for several plant- and lab-compacted mixtures across the Midwest in the United States. For further validation, sensitivity of the proposed model to the variations of the recycled material (RAP and RAS), asphalt content, binder grade type, crumb rubber content, aggregate type, gradation, and test temperature were evaluated. To investigate the effect of each factor, the factor of interest was incrementally changed while all other variables were held constant (Figure 9 to Figure 14). In order to provide an insight into the performance of the CNN model, the predictions made by a previously developed GEP model [26] was included in the analysis. A short summary is now provided.

Gene expression programming (GEP) was implemented in the development of an earlier rutting prediction model [26]. The dataset (96 data points) in the previous study was much smaller than the dataset (1000 data points) used in the current study. A smaller number of mixtures, test temperatures, and the use of only four of the available 200 points in each rutting curve were used previously. The following section compares the predictive ability of the previous GEP model to the current CNN prediction model for the Hamburg lab rut test. It is acknowledged that this is not a perfectly fair comparison, since a much larger training set was available for training of the CNN model. Nonetheless, the comparison illustrates the strides made in moving to a more powerful machine learning (ML) model, combined with the benefits of expanding the training data set as more mixture data was collected.

## 5.1. Effect of RAP/RAS

Figure 9 shows the effect of recycled content on the previous (GEP) and current (CNN) machine learning model predictions. The following conclusions can be drawn:

- The CNN model is capable of capturing differences in recycling content, where increased ABR clearly reduces rutting potential. Figure 3a shows the effects of increasing recycled content in a base mixture. It shows that even modest amounts of RAP and RAS promote rut resistance.
- Figure 9 b & c highlights the improvements of the CNN model over the GEP model by comparing the rut depth values for different passes for a mix with different levels of RAP and RAS. While the GEP model is "unsteady," the CNN model provides an expected and a steady trend of the mixture rutting performance with an increase in RAP/RAS.

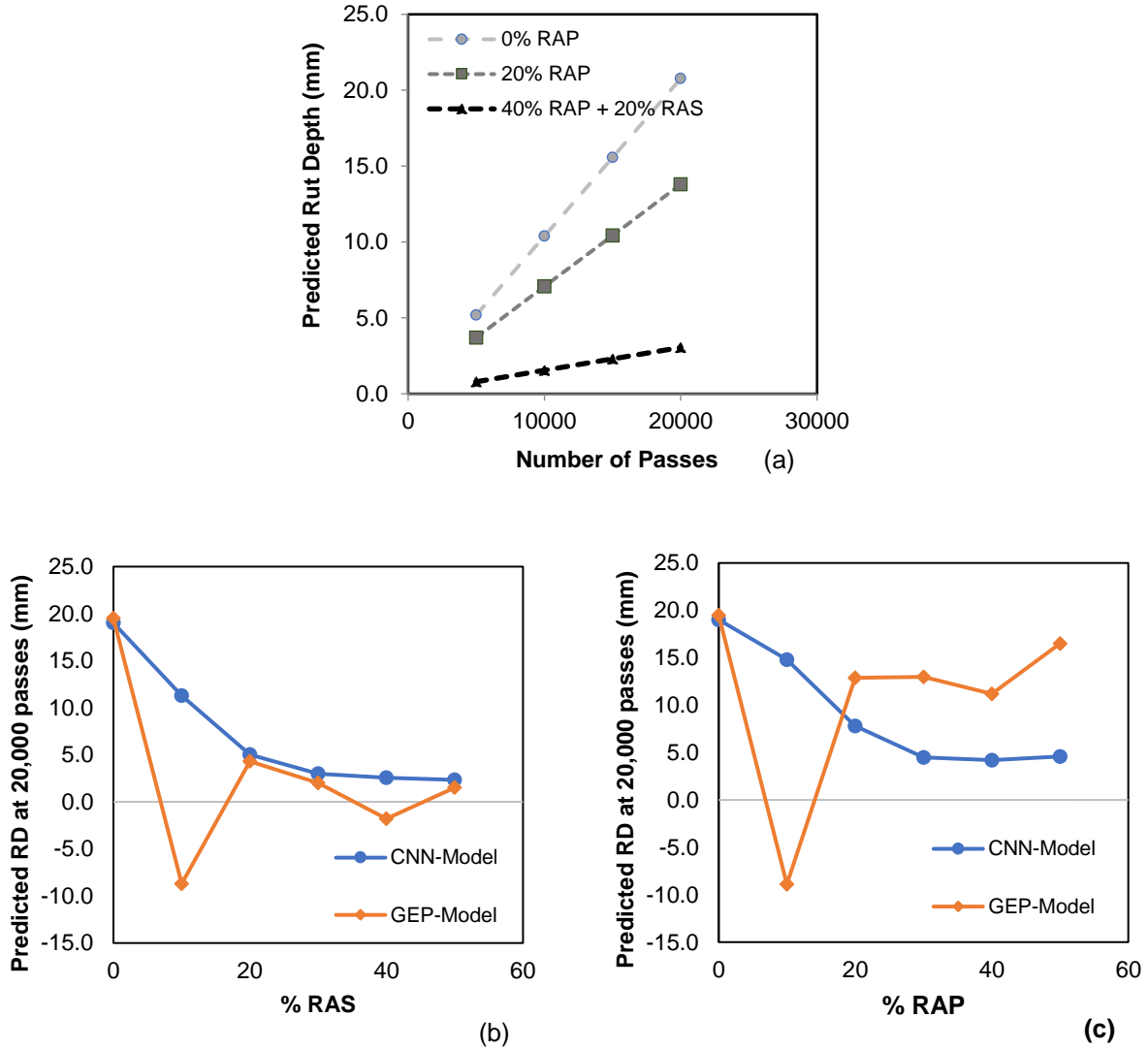

**Figure 9.** Sensitivity analysis of the CNN and GEP models: a) Effect of RAP and RAS, b) Effect of RAS and c) Effect of RAP

## 5.2. Effect of Test Temperature

Figure 10 shows model sensitivity to test temperature. The following observations were made:

- Figure 10a shows predicted rut depth for a base mixture at various passes and test temperatures by the new CNN model. The observed trend was expected as rut depth values increased at warmer testing temperatures.
- Figure 10b shows the sensitivity comparison of GEP and CNN models with respect to testing temperatures. While the GEP model was able to predict the expected overall trend, the negative rut depth prediction by the GEP model at 40 and 46°C were unexpected. .

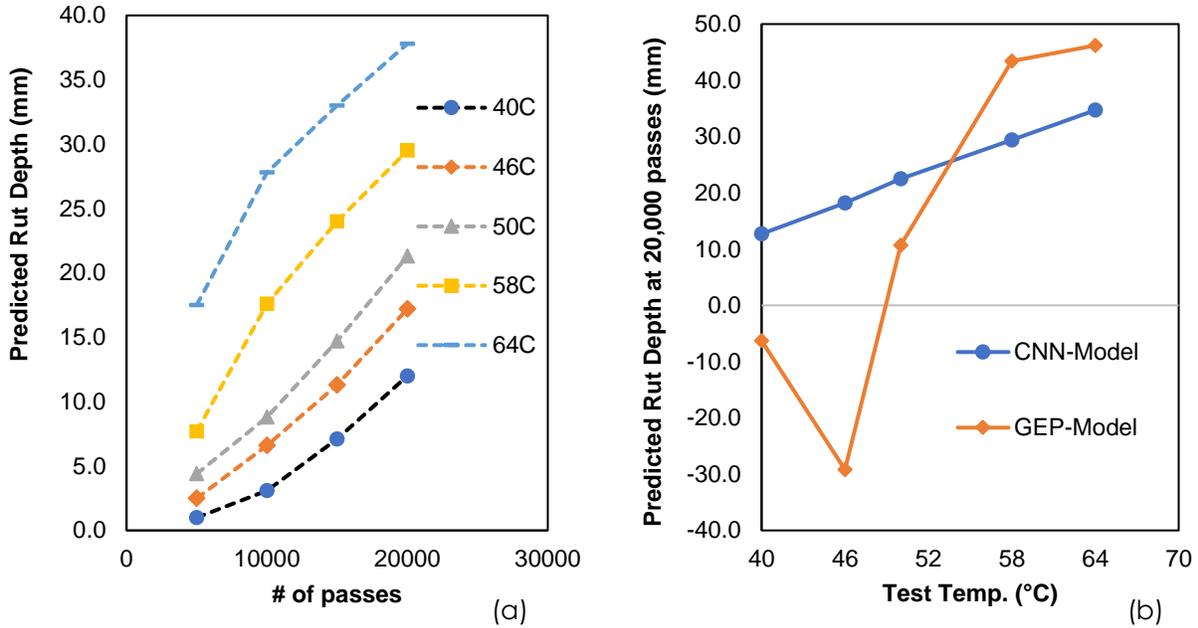

**Figure 10.** Sensitivity analysis for CNN and GEP models: a) Effect of temperature on rut depth versus number of passes in CNN model, b) Effect of test temperature on rut depth in CNN and GEP models

### 5.3. Effect of Binder PG High Temperature

Figure 11 shows CNN model sensitivity to binder PG high temperature grade.

- Figure 11a shows the rut depth prediction with binders that are progressively stiffer. As expected, the model predicts a lower rut depth with stiffer binder grade.
- Figure 11b shows the comparison of GEP and CNN models- both models are able to exhibit satisfactory trends for binder type.

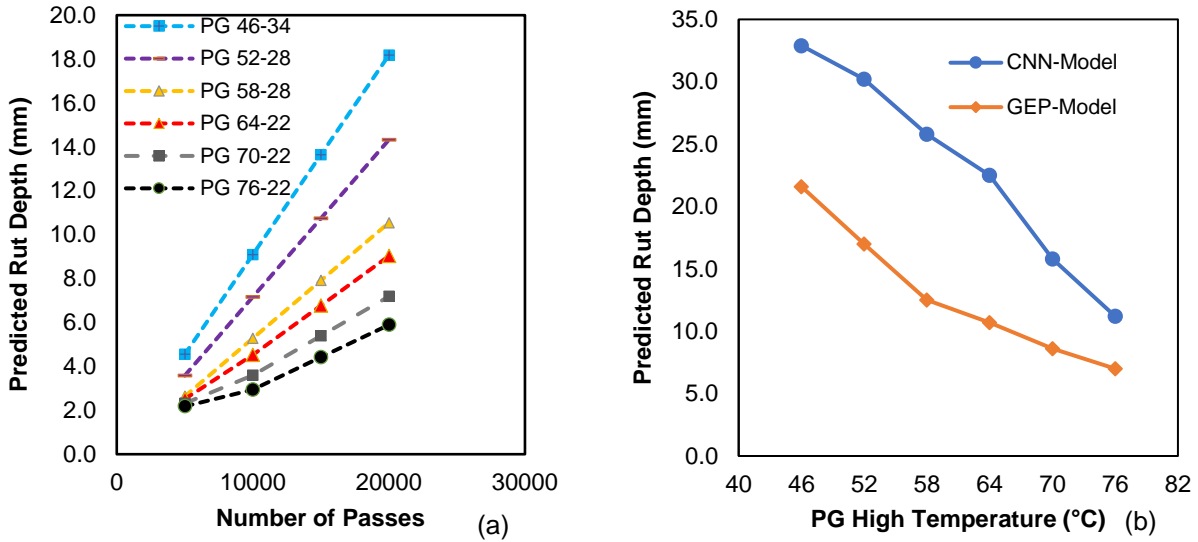

**Figure 11.** Effect of Binder PG high temperature on model prediction: (a) CNN model predictions with progressively stiffer binder, (b) Comparison of the CNN and GEP predictions with progressively stiffer binders

### 5.4. Effect of Mixture Production (Lab vs. Plant)

The current model also has the ability to distinguish between lab and plant produced mixtures, as shown in Figure 12a. It should be noted that the difference between the rutting performance of the plant- and the lab-produced mixture is very subtle in the case shown here, but this could differ with other mixtures. For the purposes of this study, it was deemed enough to show that the model can be developed to be sensitive to mixture production differences. As always, the more robust the training data set, the more robust the future prediction capability of the ML model.

Figure 12b shows GEP model sensitivity to plant- and lab-produced mixtures. While the model is able to differentiate between the mixtures, the differences are glaringly large and unexplainable. For example, the GEP model predicts 4 mm rut depth across all passes for the lab mix.

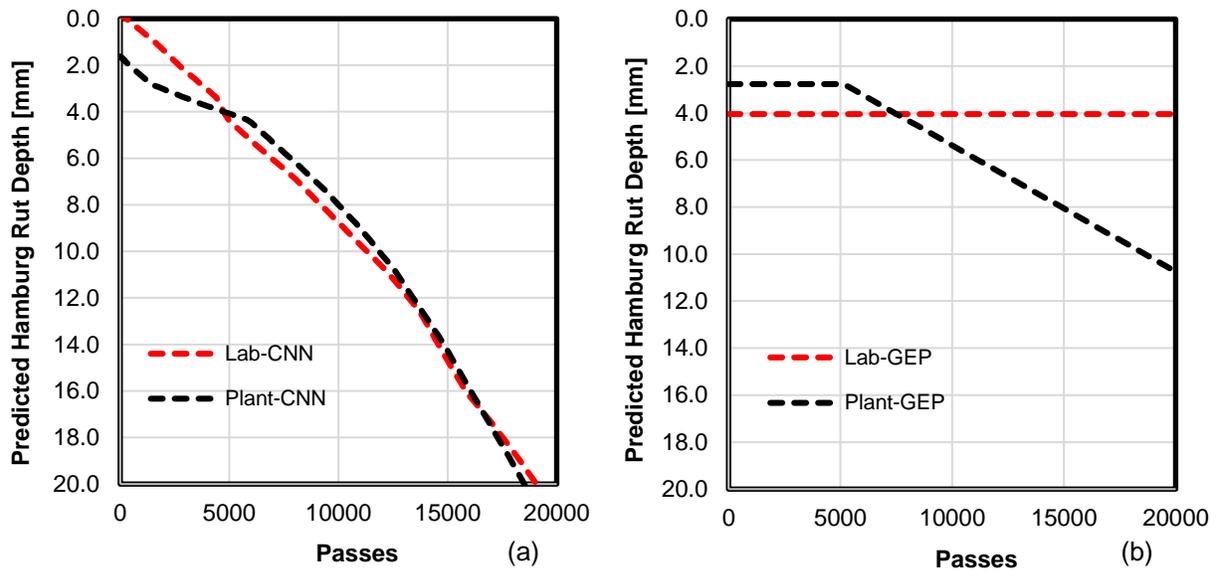

**Figure 12.** Effect of production (Lab vs. Plant produced) on model predictions: (a) CNN model, (b) GEP model

## 5.5. Effect of Crumb Rubber

Figure 13 shows the effect of GTR in asphalt mixture rutting performance. As observed in Figure 13, the CNN model predicts a decrease in rutting with the addition of GTR, as expected. In this case, there was insufficient data available on GTR mixes during the training of the GEP model, so only the CNN model results have been presented in this section.

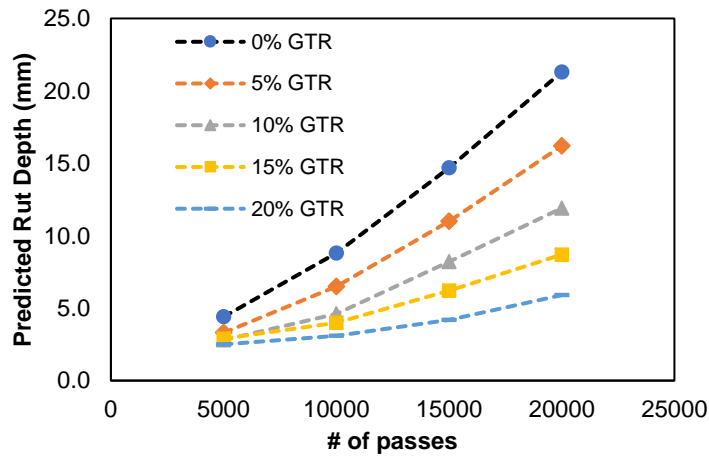

**Figure 13.** Effect of GTR% on Hamburg rutting curve as predicted by CNN model

## 5.6. Effect of aggregate type/gradation

Currently, the CNN model has been trained using only two types of aggregate- limestone and granite, and two types of gradations- dense-graded and stone matrix (mastic) asphalt (SMA). Furthermore, due to the limitation of available data, the dense-graded mixes could only be assigned to have limestone as an aggregate, while the SMAs were assigned granite as aggregate. Keeping these limitations in mind, Figure 14 shows the trends exhibited by the CNN model prediction for the two aggregate types and gradation with respect to changing mixture constituents. As seen in both cases, the trends are similar. The base mixture with 64-22 binder (6.6% total asphalt content) and 0% RAP shows high rutting potential. Addition of 45% RAP to the next mix iteration (with 6.2% total binder content) results in better rutting resistance. Further, switching the PG64-22 with to a softer base binder, PG 58-28 and subsequently PG46-34, causes progressively higher rutting in the mix.

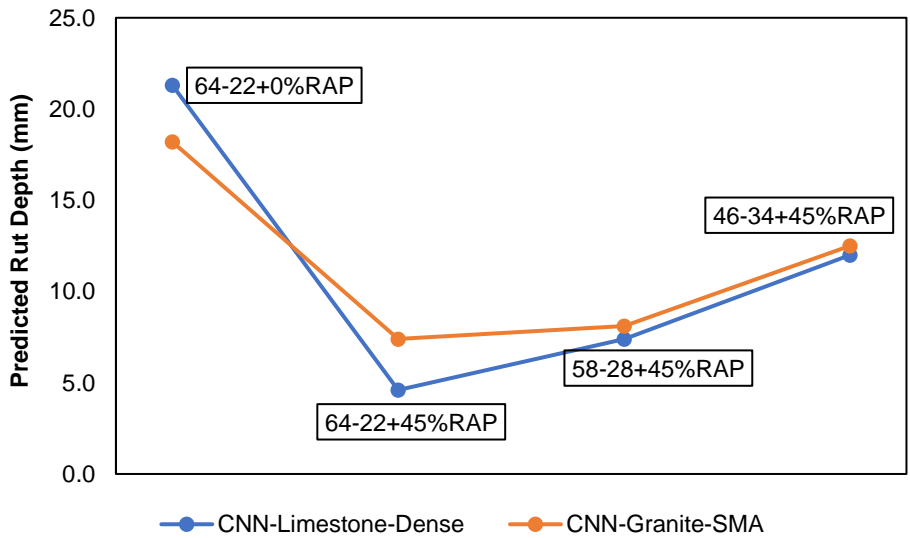

**Figure 14.** The CNN model predictions for aggregate type and gradation

## 6. PERFORMANCE-SPACE DIAGRAM TOOL

Majidifard et al. [26], [27] implemented the GEP-based models to predict the low- and high-temperature performance in terms of DC(T) fracture energy and Hamburg rut depth, respectively. The authors used a concept called Performance-Space diagram (PSD), introduced by Buttlar et al. [50], to show the efficiency of the tool in performance prediction. PSD allows a holistic view of a mixture performance, which helps in implementing a balanced mix design approach. For example, Rath et al. [51] used PSD to show the possibility of using higher recycling in asphalt mixtures if the introduced stiffness due to recycled material was "balanced" with use of a softer base binder. A database containing a comprehensive collection of Hamburg and DC(T) tests results were used to develop these machine learning-based prediction models. The models were formulated in terms of typical influencing mixture properties variables such as asphalt binder high-temperature performance grade (PG), mixture type, aggregate size, aggregate gradation, asphalt content, total asphalt binder recycling content and tests parameters like temperature and number of cycles (Figure 15). The model's accuracy was assessed through a rigorous validation process and found to be quite acceptable, despite the relatively small size of the training set. Figure 15 shows the schematic view of the software. After filling the inputs on the left and clicking "Run" button, the tool will populate the Hamburg-DC(T) PSD, representing the mixture's DC(T) fracture energy at specified test temperature, and rut depth at 20,000 passes (10,000 cycles) at specified test temperature.

The prediction models combined with the PSD can be used in pre-design stages of an asphalt mixture to tweak mixture designs for suitable performance in terms of cracking and rutting, helping reduce the overall cost of mixture testing in design phase. For example, data in Figure 14 shows that the rut depth was improved with the addition of RAP to the PG 64-22 binder. However, the addition of RAP increases the risk of thermal cracking. This can be remedied by switching to a softer base binder, but that would result in an increase in the rutting susceptibility. Thus, an iterative process can be applied according to the performance limits of the prevailing agency to design an optimized or 'balanced' mix in terms of cracking and rutting performance. These time-consuming iterative steps in real world can be simulated in this tool, and a balanced mixture can be made with minimum efforts and much lower expense [52].

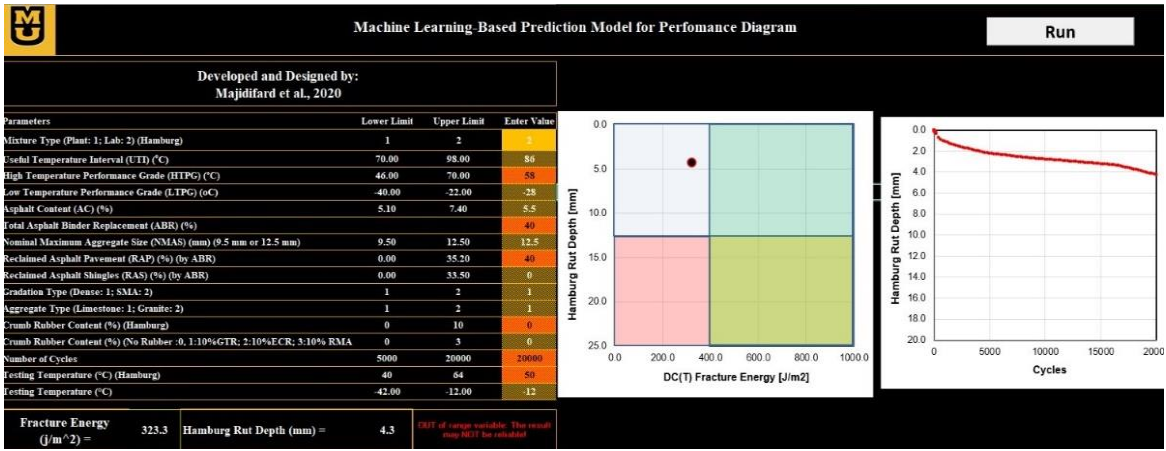

**Figure 15.** The GEP (fracture energy) – CNN (rutting) based prediction model for performance space diagram developed by Majidifard et al. [27], [52].

## 7. COMPARISON OF RUT PROFILES OBTAINED FROM THE CNN AND GEP MODELS

This section investigates the trends seen in the predicted rut depth values as obtained by CNN and GEP models to highlight the substantial improvements provided by the CNN model over the GEP model. Selected rut depth profiles from the investigated mixtures were plotted as predicted using the GEP- and CNN-based models (Figure 16). The following conclusions were then drawn:

a. The rut depth curve prediction in the GEP model was based on linear interpolation between the values predicted at discrete passes (four data points from each curve), such as 5000, 10000, 15000, and 20000 passes. This results in a rather linear behavior of the predicted rut depth profile by the GEP model. The CNN model, however, was based on data from 200 data points during 20,000 passes, allowing the model to improve its accuracy in profile prediction (Figure 16).

b. Due to the use of rut depth values at discrete passes, the accuracy of rut depth prediction at an intermediate number of wheel-pass, say 7500 passes, is lower than the average accuracy of the model. Using all points through 20,000 passes in the CNN model enables the user to accurately predict rut depth at any number of passes. This feature makes the model highly versatile and applicable to various cases where the number of passes is adjusted according to the requirements of the project (climates and traffic).

c. It was observed that certain asphalt mixes with recycled materials, especially roofing shingles, resulted in negative rut depth with the old GEP model. Figure 16 (c) and (d) show such a trend when presence of recycled roofing shingles in the mixtures lead to a negative constant rut depth profile in the GEP model. This was due to the lack of a broad range of data points available for recycled mixtures. The CNN model provides an improved prediction capability for recycled mixtures, as shown in Figure 16. It is noted by the authors that even higher accuracy can be expected in future iterations of this tool with a larger training dataset for recycled mixtures.

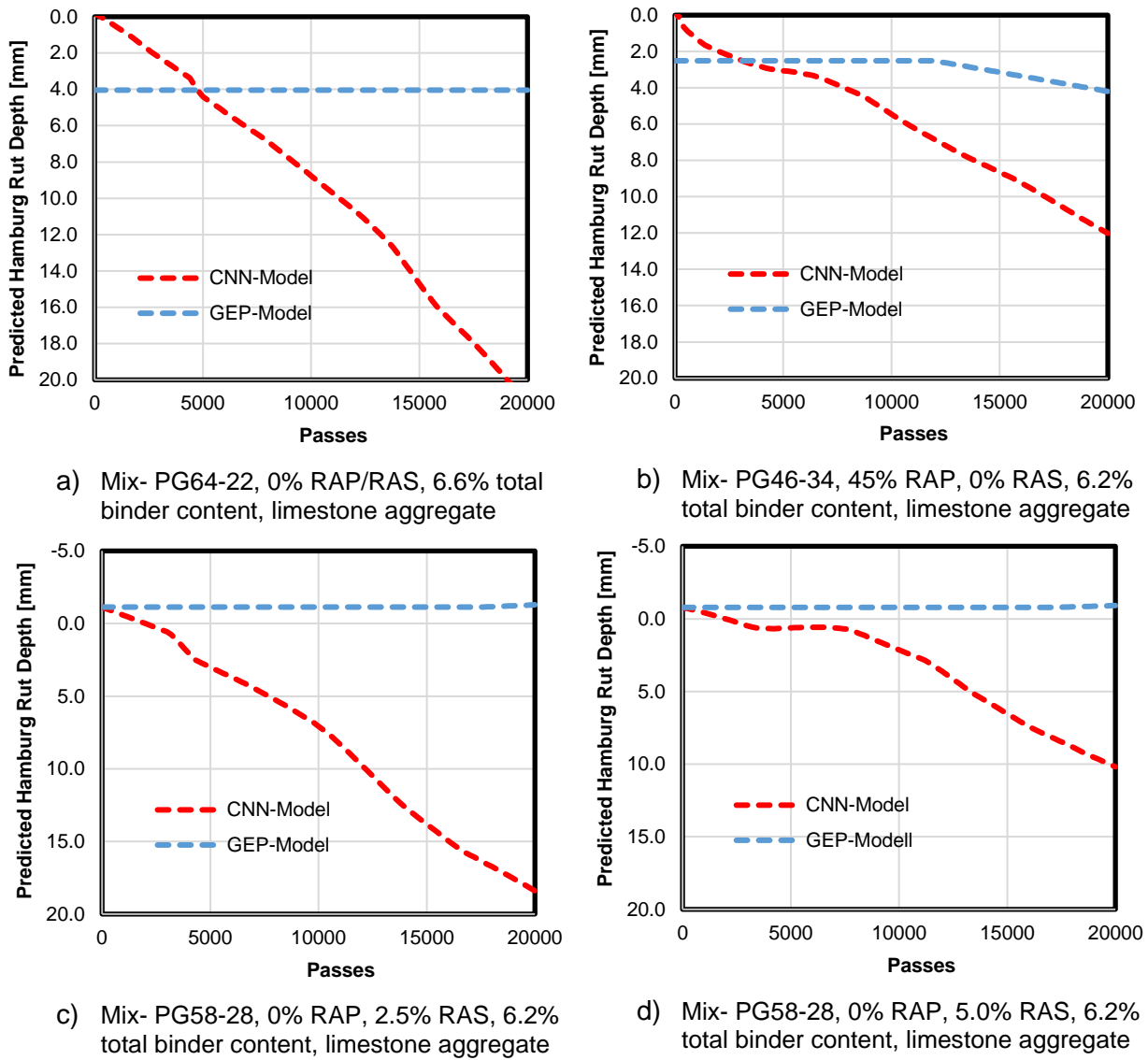

a) Mix- PG64-22, 0% RAP/RAS, 6.6% total binder content, limestone aggregate

b) Mix- PG46-34, 45% RAP, 0% RAS, 6.2% total binder content, limestone aggregate

c) Mix- PG58-28, 0% RAP, 2.5% RAS, 6.2% total binder content, limestone aggregate

d) Mix- PG58-28, 0% RAP, 5.0% RAS, 6.2% total binder content, limestone aggregate

**Figure 16.** Comparison of the rut depth profiles predicted by the GEP and CNN models.

## 7.1. Full Hamburg Curves Using CNN Model for Plant-Produced Mixtures

Full Hamburg rutting curves of PPLC mixtures from the states of Missouri and Illinois were plotted to compare measured vs. predicted rut depth profiles. Details about the mixtures can be found elsewhere [23], [41], [53], but brief details of the mixtures are shown in Table . Figure 17 shows the rut depth profile measured in the Hamburg device compared to the predicted rut depth by the CNN and GEP models. As can be observed, the CNN model was able to increase the accuracy of the prediction of rut depth substantially in comparison to the GEP model. The CNN model was accurately able to predict the entire rut depth profile at different temperatures.

**Table 3.** Mixture Details

| Mix ID | Binder PG used (%) | NMAS (mm) | ABR% by RAP | ABR% by RAS | Crumb Rubber % |
|---|---|---|---|---|---|
| MO13_1 | 70-22 (5.7) | 9.5 | 17 | 0 | 0 |
| US54_1 | 58-28 (5.2) | 12.5 | 0 | 33 | 0 |
| 1807 | 46-34 (6.2) | 19 | 34.4 | 14.0 | 0 |
| 1818 | 64-22 (5.7) | 9.5 | 20.4 | 0 | 0 |
| 1829 | 70-28 (7.9) | 4.75 | 17.8 | 9.3 | 10 |
| 1835 | 46-34 (5.9) | 12.5 | 25.0 | 16.1 | 10 |

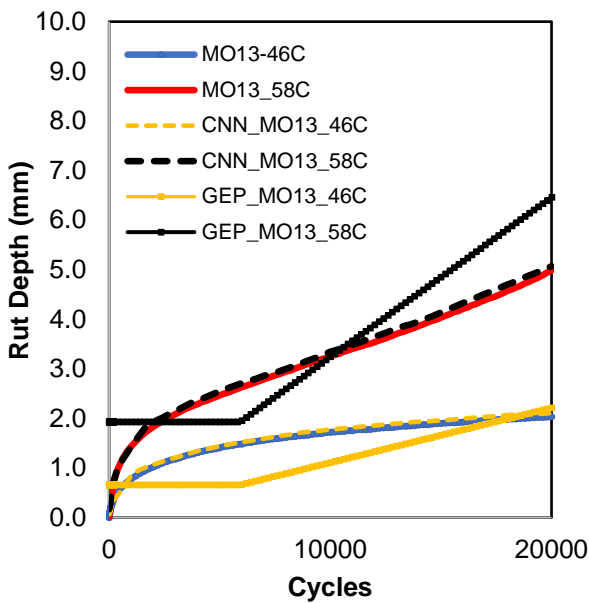
a)

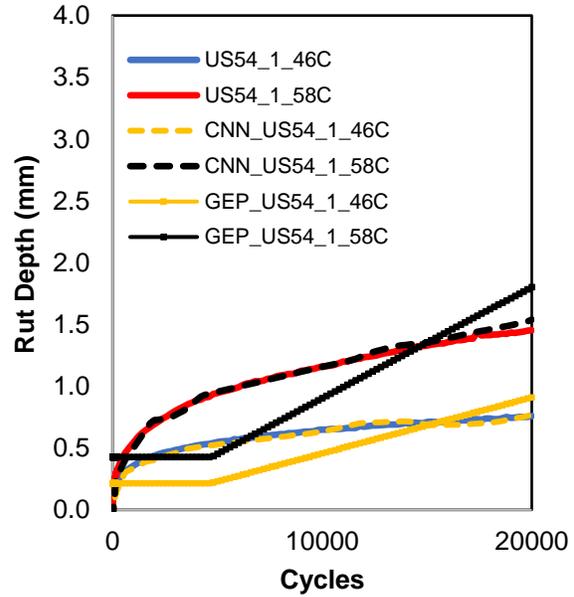
b)

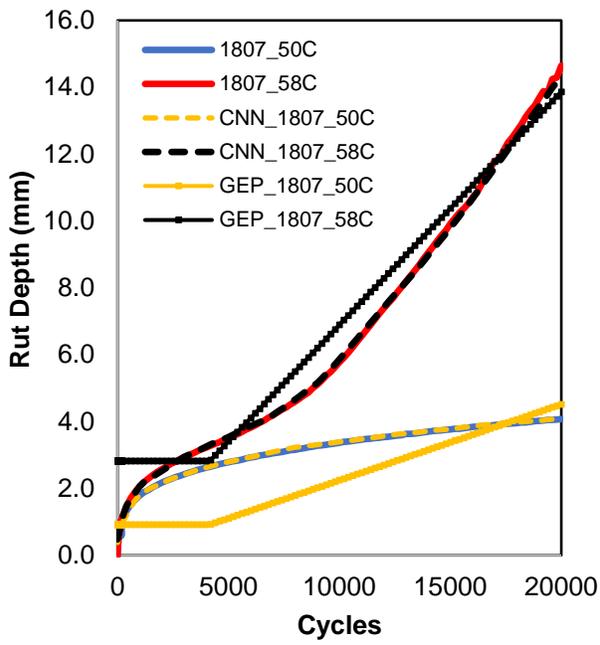
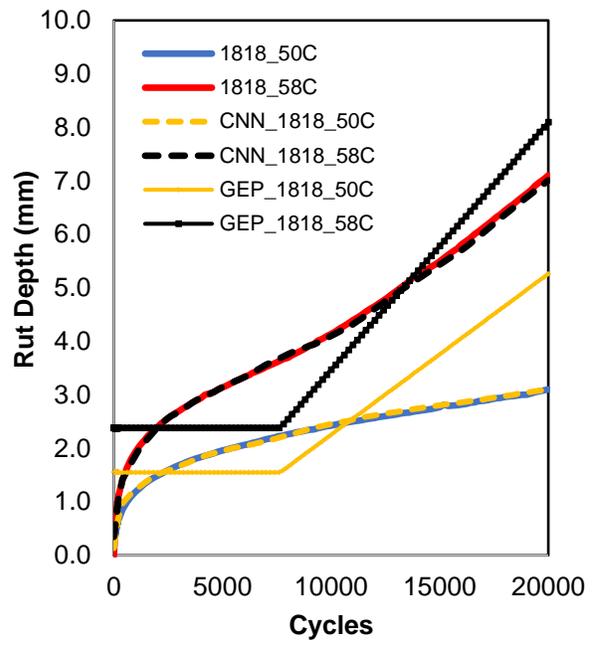

(c) (d)

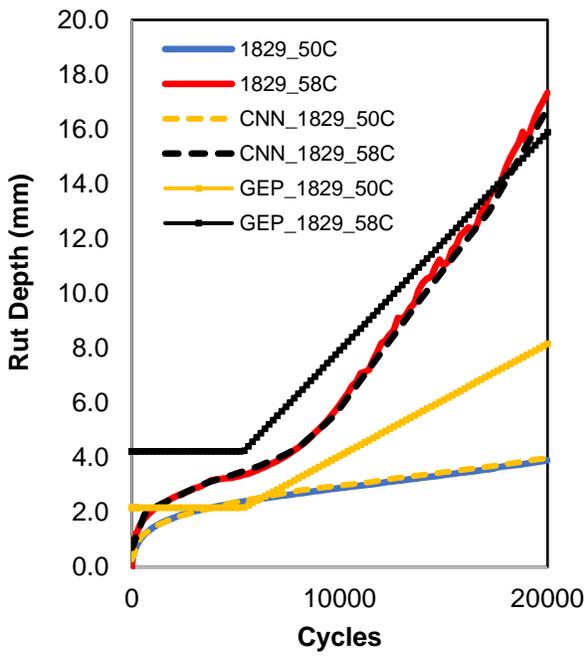
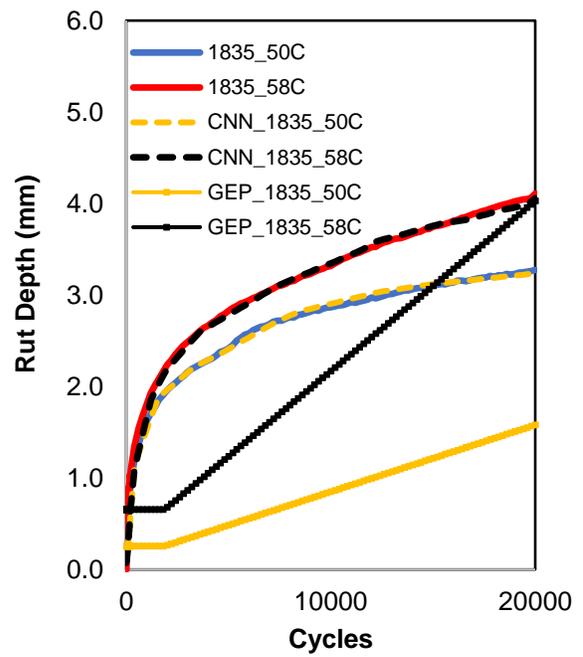

(e) (f)

**Figure 17.** Comparison of predicted and actual Hamburg rut depth using the GEP- and CNN models

## 8. CONCLUSION

In this paper, a new model was developed to predict the rutting depth of asphalt mixtures using a deep learning framework, namely CNN. The dataset contained 10,000 data points collected from 50 mixture test results. Thirteen mixture parameters and testing variables were considered to develop the model. The results confirmed the acceptable performance of CNN algorithm. The results of sensitivity analysis showed that the proposed model is highly sensitive to RAP, RAS, high PG, crumb rubber content, test temperature, and numbers of cycles. The proposed model predicts the full rutting curve of asphalt mixtures across 20,000 passes at various test temperatures in the HWTT. Furthermore, a CNN-based prediction model for performance space diagram outputs is presented as a tool to predict both low- and high-temperature mixture performance in terms of DC(T) fracture energy and Hamburg rut depth. The prediction models can be used in pre-design stages of an asphalt mixture to tweak mixture designs for suitable performance in terms of cracking and rutting, helping reduce the overall cost of mixture testing in the design phase. A comparative study has been performed between a GEP-based prediction model and the new CNN model. The results clearly demonstrate that the CNN model performance is superior to the GEP model. It should also be noted that by increasing the dataset's size, CNN models operate better than classical models like GEP. On the other hand, classic algorithms may perform more accurately when the dataset's size is small. The proposed model can be updated with additional data in the future to provide more accurate predictions over a wider range of mixture component materials, such as aggregate mineralogy, shape, and gradation. More HWTT results with a wider range of test temperatures will also be incorporated into future model development to assist agencies in evaluating proper HWTT test temperatures for the purpose of balanced mix design in various climates.

**ACKNOWLEDGEMENTS AND DISCLAIMER**

The research team would like to, first and foremost, thank the Missouri Department of Transportation (MoDOT) and Missouri Asphalt Pavement Association (MAPA) for their generous support of this research. The authors would also like to thank colleagues at the Illinois Tollway. Finally, the research team would like to thank James Meister, and Grant Nichols from Mizzou Asphalt Pavement and Innovation Laboratory (MAPIL) for their assistance in during the course of this study.

VA, 2019, pp. 39–50.

[12] D. Batioja-Alvarez, J. Lee, R. Rahbar-Rastegar, and J. E. Haddock, "Asphalt Mixture Quality Acceptance using the Hamburg Wheel-Tracking Test," *Transp. Res. Rec. J. Transp. Res. Board*, p. 036119812091974, 2020, doi: 10.1177/0361198120919749.

[13] E. Mahmoud, E. Masad, S. Nazarian, and I. Abdallah, "Modeling and Experimental Evaluation of Influence of Aggregate Blending on Asphalt Mixture Strength," pp. 48–57, doi: 10.3141/2180-06.

[14] F. Yin, E. Arambula, R. Lytton, A. E. Martin, and L. G. Cucalon, "Novel method for moisture susceptibility and rutting evaluation using Hamburg wheel tracking test," *Transp. Res. Rec.*, vol. 2446, pp. 1–7, 2014, doi: 10.3141/2446-01.

[15] N. Tran, A. Taylor, P. Turner, C. Holmes, and L. Porot, "Effect of rejuvenator on performance characteristics of high RAP mixture," *Road Mater. Pavement Des.*, vol. 18, no. February, pp. 183–208, 2017, doi: 10.1080/14680629.2016.1266757.

[16] H. Ozer, I. L. Al-Qadi, A. I. Kanaan, and D. L. Lippert, "Performance Characterization of Asphalt Mixtures at High Asphalt Binder Replacement with Recycled Asphalt Shingles," *Transp. Res. Rec. J. Transp. Res. Board*, vol. 2371, no. 1, pp. 105–112, 2013, doi: 10.3141/2371-12.

[17] H. Majidifard, N. Tabatabaee, and W. Buttlar, "Investigating short-term and long-term binder performance of high-RAP mixtures containing waste cooking oil," *J. Traffic Transp. Eng. (English Ed.*, vol. 6, no. 4, pp. 396–406, 2019.

[18] G. H. Hamedi, A. Sahraei, and M. R. Esmaeeli, "Investigate the effect of using polymeric anti-stripping additives on moisture damage of hot mix asphalt," *Eur. J. Environ. Civ. Eng.*, pp. 1–14, 2018.

[19] W. Buttlar, P. Rath, H. Majidifard, E. V Dave, and H. Wang, "Relating DC (T) Fracture Energy to Field Cracking Observations and Recommended Specification Thresholds for Performance-Engineered Mix Design," *Asph. Mix.*, vol. 51, 2018.

[20] A. Ghanbari, B. S. Underwood, and Y. R. Kim, "Development of a rutting index parameter based on the stress sweep rutting test and permanent deformation shift model," *Int. J. Pavement Eng.*, pp. 1–13, 2020.

[21] Y. D. Wang, A. Ghanbari, B. S. Underwood, and Y. R. Kim, "Development of a performance-volumetric relationship for asphalt mixtures," *Transp. Res. Rec.*, vol. 2673, no. 6, pp. 416–430, 2019.

[22] Y. R. Kim, C. Castorena, Y. Wang, A. Ghanbari, and J. Jeong, "Comparing Performance of Full-depth Asphalt Pavements and Aggregate Base Pavements in NC," FHWA/NC/2015-02 Report, 2018.

[23] B. Jahangiri, H. Majidifard, J. Meister, and W. G. Buttlar, "Performance Evaluation of Asphalt Mixtures with Reclaimed Asphalt Pavement and Recycled Asphalt Shingles in Missouri," *Transp. Res. Rec.*, 2019, doi: 10.1177/0361198119825638.

[24] A. Buss, R. C. Williams, and S. Schram, "Evaluation of moisture susceptibility tests for warm mix asphalts," *Constr. Build. Mater.*, vol. 102, pp. 358–366, 2016, doi: 10.1016/j.conbuildmat.2015.11.010.

[25] B. Jahangiri, P. Rath, H. Majidifard, L. Urra, and W. G. Buttlar, "OnlyA Comprehensive Performance Investigation of Asphalt Mixtures in Missouri: Laboratory, Field, and ILLI-TC Modeling," *Road Mater. Pavement Des.*

[26] H. Majidifard, B. Jahangiri, P. Rath, W. G. Buttlar, and A. H. Alavi, "Developing a Prediction Model for Rutting Depth of Asphalt Mixtures Using Gene Expression
31